\documentclass[conference]{IEEEtran}
\usepackage{url}
\usepackage{cite}
\usepackage{amsmath,amssymb,amsfonts}
\usepackage{algorithmic}
\usepackage{graphicx}
\usepackage{textcomp}
\usepackage{xcolor}
\usepackage{booktabs}
\usepackage{multirow} 
\usepackage{promptbox}

\def\BibTeX{{\rm B\kern-.05em{\sc i\kern-.025em b}\kern-.08em
    T\kern-.1667em\lower.7ex\hbox{E}\kern-.125emX}}
\begin{document}

\title{RooseBERT: A New Deal For Political Language Modelling}
\author{\IEEEauthorblockN{Deborah Dore}
\IEEEauthorblockA{\textit{Université Côte d’Azur}\\ \textit{CNRS, INRIA, I3S}\\
Sophia Antipolis, France \\
deborah.dore@univ-cotedazur.fr}
\and
\IEEEauthorblockN{Elena Cabrio}
\IEEEauthorblockA{\textit{Université Côte d’Azur}\\ \textit{CNRS, INRIA, I3S}\\
Sophia Antipolis, France \\
elena.cabrio@univ-cotedazur.fr}
\and
\IEEEauthorblockN{Serena Villata}
\IEEEauthorblockA{\textit{Université Côte d’Azur}\\ \textit{CNRS, INRIA, I3S}\\
Sophia Antipolis, France \\
serena.villata@cnrs.fr}
}

\maketitle

\begin{abstract}
The increasing amount of political debates and politics-related discussions calls for the definition of novel computational methods to automatically analyse such content with the final goal of lightening up political deliberation to citizens. However, the specificity of the political language and the argumentative form of these debates (employing hidden communication strategies and leveraging implicit arguments) make this task very challenging, even for current general-purpose pre-trained Language Models (PLMs). To address this, we introduce a novel PLM for political discourse language called RooseBERT. Pre-training a language model on a specialised domain presents different technical and linguistic challenges, requiring extensive computational resources and large-scale data. RooseBERT has been trained on large political debate and speech corpora (11GB) in English. To evaluate its performances, we fine-tuned it on multiple downstream tasks related to political debate analysis, i.e., stance detection, sentiment analysis, argument component detection and classification, argument relation prediction and classification, policy classification, named entity recognition (NER). 
Our results show improvements over general-purpose PLMs on the majority of these tasks, highlighting how domain-specific pre-training enhances performance in political debate analysis. We release RooseBERT for the research community\footnote{\url{https://huggingface.co/collections/MARIANNE-INRIA/roosebert}}.
\end{abstract}


\section{Introduction}
\label{sec:introduction}
Political debates, politics-related (online) discussions and opinions represent key elements of public discourse and deliberation, and they are fundamental for citizens to take informed decisions. The ability to automatically analyse such a kind of content is therefore central in Artificial Intelligence, where different aspects of political discourse have been investigated, e.g., performance evaluation~\cite{liu2024potus}, argumentative structures~\cite{HaddadanCV19}, sentiment analysis~\cite{abercrombie2020parlvote}, and logical fallacies~\cite{GoffredoHVCV22}. 
These approaches rely on pre-trained Language Models (PLMs) originally developed for general-purpose corpora, e.g., BERT~\cite{devlin-etal-2019-bert}. Although these models achieve good results in some of these tasks, they underperform when the goal is to capture the unique features of political content (e.g., domain-specific terminology, implicit argumentation, strategic communication patterns). In the literature, researchers have shown that domain-specific variations in vocabulary and semantics can significantly affect model performance~\cite{nooralahzadeh-etal-2018-evaluation}, avoiding vocabulary mismatches and encoding semantic shifts~\cite{lee2020biobert}.




To address this issue, we introduce RooseBERT, a domain-specific LM pre-trained on English political debates. Built on the BERT architecture, widely adopted for domain adaptation tasks~\cite{beltagy-etal-2019-scibert, huang2019clinicalbert, lee2020biobert, yang2020finbert, kawintiranon-singh-2022-polibertweet, hu-etal-2022-conflibert, WadaTOMKKM24, ZEINALI2024190} and empirically selected over larger architectures which overfit on our domain corpus, RooseBERT is tailored to capture the distinctive features of political debates. Our aim is to improve the performance of state-of-the-art models on tasks such as sentiment analysis, stance detection, and argument mining which are relevant to political analysis. Encoder-based architectures are especially well-suited for such tasks due to their strength in encoding semantic relationships and contextual information, showing strong performance on a range of downstream tasks relevant to political discourse analysis~\cite{abercrombie2020parlvote, liu2021aaai, GoffredoCVHS23}. 

\noindent Our main contributions are:
\begin{itemize}
    \item We train and release RooseBERT, alongside the largest publicly available English political debate corpus to date (11GB, spanning 1946–2025 across 11 geopolitical contexts)\footnote{\url{https://huggingface.co/datasets/MARIANNE-INRIA/PoliticalDebatesCorpus-EN}}. 
    \item We extensively evaluate the model across multiple downstream tasks using ten different datasets for political discourse analysis: Stance Detection, Sentiment Analysis, Argument Component Detection and Classification, Argument Relation Prediction and Classification, Policy Classification, and NER. RooseBERT achieves better performance on 8 out of 10 tasks.
    \item We demonstrate that domain-adaptive pre-training on a BERT-scale architecture achieves competitive or superior performance to ModernBERT, a larger, newer general-purpose model, while requiring only 24 hours of training on 8 A100 GPUs and fine-tuning on a single GPU, providing a lightweight solution for NLP tasks on political data. RooseBERT generalises well across different types of political debates from various geopolitical contexts, including televised U.S. presidential debates, U.K. and South Africa parliamentary discussions.
\end{itemize}


\section{Related Work}
\label{sec:related_work}
Word embeddings are the foundation of most NLP systems, encoding the semantic and syntactic properties that downstream tasks rely on.
%
BERT~\cite{devlin-etal-2019-bert} leverages a transformer architecture to produce highly context-sensitive embeddings. It was trained using two different objective learning functions, Masked Language Modelling (MLM) and Next Sentence Prediction (NSP), with the ultimate goal of learning bidirectional contextual representations and capturing sentence-level relationships.
BERT was trained on a large general domain corpora (BooksCorpus and English Wikipedia) and it struggles to capture semantics related to specialised domains. To address this, models have been proposed by pre-training BERT on specific domains, e.g. ClinicalBERT for medical applications~\cite{huang2019clinicalbert}, SciBERT for scientific texts~\cite{beltagy-etal-2019-scibert}, BioBERT for the biomedical domain~\cite{lee2020biobert}, FinBERT for finance~\cite{yang2020finbert}, BERTweet for Twitter data~\cite{nguyen-etal-2020-bertweet} and Mol-BERT~\cite{li2021molbert} for molecular representation.

In the political domain, well-constructed embeddings are particularly important for capturing domain-specific terminology.
Some PLMs have been proposed: \cite{hu-etal-2022-conflibert} introduced ConfliBERT for conflict and political violence. It has been trained using 12 datasets and has been evaluated on 18 tasks showing that the improvement of ConfliBERT is statistically significant in all tasks but three; 
\cite{kawintiranon-singh-2022-polibertweet} presented PoliBERTweet, a PLM trained from BERTweet~\cite{nguyen-etal-2020-bertweet} on 83M US 2020 election-related English tweets, outperforming TweetEval~\cite{BarbieriCAN20} and BERTweet~\cite{nguyen-etal-2020-bertweet} on all tasks related to the political domain.
While ConfliBERT and PoliBERTweet are valuable political-domain models, they are not tailored to the dialogic and argumentative nature of political debates. 
Both models lack the turn-taking, rebuttals, and rhetorical complexity central to debates.

This motivates the introduction of RooseBERT, which to the best of our knowledge, is the first PLM specifically trained on full debate transcripts (presidential and parliamentary), capturing cross-speaker interactions, persuasive framing, and argument structures. Its debate-specific corpus and cross-national coverage enable stronger performance on tasks like argument mining and sentiment analysis, where the debate context is crucial.

    \section{Methodology}
\label{sec:methodology}
In this section, we first describe the datasets used for pre-training RooseBERT, then we present our pre-training strategy, the description of the downstream tasks, and the evaluation metrics.

\subsection{Corpora}
To pre-train RooseBERT on English political debates, we assembled a comprehensive corpus of debate transcripts and related material totalling 11GB, which we release publicly as a standalone contribution. To the best of our knowledge, this is the largest open English political debate corpus available, and it is designed to support future work in political NLP. Our complete dataset combines existing curated collections and transcripts scraped from official debate websites, ensuring broad coverage of formats (televised exchanges, parliamentary debates, primary debates) over a comprehensive time period (1946–2025), capturing the evolution of rhetorical strategies and issue framing across decades and geopolitical contexts. All datasets originate from authoritative speakers and official political settings (i.e., presidential debates and parliamentary sessions), where political decisions are discussed, contested, and communicated to the public. 

Each dataset was pre-processed to remove hyperlinks, markup tags and to collapse multiple spaces. We report the size of each dataset:

\noindent \textbf{African Parliamentary Debates.}
The HOME Project Parliamentary Activity Dataset (573MB) covers the Ghanaian and the South African Parliament from 1999 to 2024~\cite{HomeDataset2024}.

\noindent \textbf{Australian Parliamentary Debates.}
It contains the debates (1GB) from each sitting day in the Australian Parliament from 1998 to 2025~\cite{AusHansard2025}.

\noindent \textbf{Canadian Parliamentary Debates.}
They were collected from the official OpenParliament website~\footnote{\url{https://openparliament.ca/debates/}} which hosts a comprehensive archive from 1994 to 2025 (1.1GB of data).

\noindent \textbf{European Parliamentary Debates.}
The EUSpeech~\cite{schumacher2016Euparl} (110MB) is a dataset of 18,403 speeches from EU leaders (heads of government in 10 member states, EU commissioners, EU Parliament, EU Central Bank and International Monetary Fund) from 2007 to 2015.

\noindent \textbf{Irish Parliamentary Debates.} 
The ParlEE Plenary Speeches dataset~\cite{sylvester2022parlee} contains, among other sources, parliamentary speeches from the Irish Parliament covering the period 2009–2019. In addition, we used the Database of Parliamentary Speeches in Ireland~\cite{IrishParliament2017}, which provides a complete record of debates from Dáil Éireann (the principal chamber of the IE Parliament) from 1919 to 2013. Together, these sources amount to approximately 3.4 GB of parliamentary debate data.

\noindent \textbf{New Zealand Parliamentary Debates.}
The ParlSpeech~\cite{NewZealand2020} dataset (791MB) contains New Zealand's parliamentary speeches covering the period from 1987 to 2019.

\noindent \textbf{Scottish Parliamentary Debates.}
The ParlScot corpus~\cite{Braby2021parlscot} (443MB) contains full-text vectors for 1.8 million spoken contributions for the Scottish Parliament until 2021.

\noindent \textbf{United Kingdom Parliamentary Debates.} 
The House of Commons Parliamentary Debates~\cite{blumenau2021house} (2.6GB) contains data on all parliamentary debates held in the House of Commons between 1979 and 2019. The data includes over 2.5 million speeches drawn from over 50 thousand parliamentary debates.

\noindent \textbf{United Nations Debates.}
The United Nations General Debate Corpus (UNGDC) is a collection of the UN General Assembly debates (1946-2023) with over 10k speeches from 202 countries ~\cite{ungdc} (186M). In addition, the UN Security Council Debates (UNSC) is a dataset of UN Security Council debates from 1992 to 2023. The corpus contains 106,302 speeches extracted from 6,233 meeting protocols (387M).

\noindent \textbf{United States Debates.}
\textit{The American Presidency Project}~\footnote{\url{www.presidency.ucsb.edu}} is an authoritative on-line source for presidential public documents. The website includes a list of presidential candidate debates from 1960-2024, including primary and general elections political debates (16M).


\subsection{Pre-training Approaches}
To effectively pre-train a LM, different approaches can be used~\cite{chalkidis2020LegalBERT, hu-etal-2022-conflibert}. One approach is to continue the pre-training \textsc{(cont)} of the original model (BERT) by initialising the model with the original weights and vocabulary, and training for additional steps on the domain-specific corpus. The second approach \textsc{(scr)} consists in training BERT from scratch, starting with a random initialisation of its weights and creating a custom vocabulary using the domain-specific corpus.
RooseBERT was trained with both approaches (\textsc{cont} and \textsc{scr}), using \textit{cased} and \textit{uncased} vocabularies.
When training the \textsc{scr} version, a custom \textit{WordPiece Tokenizer} trained on the same training corpus of English political debates was used. 

The advantage of a \textsc{scr} model is the use of a tokenizer with a domain-specific vocabulary, which can encode entire specialised terms as single tokens~\cite{beltagy-etal-2019-scibert, gu2021domain}. In contrast, tokenizers based on general-purpose vocabularies tend to break these terms into multiple sub-tokens, potentially weakening the model’s understanding of domain-specific content. For example, \textit{RooseBERT-scr-uncased}'s vocabulary shares only 56.32\% of its tokens with \textit{bert-base-uncased}’s vocabulary. Words such as \textit{deterrent}, \textit{endorse}, and \textit{bureaucrat} are represented as single tokens in RooseBERT's custom vocabulary, while they are absent from BERT’s vocabulary and, thus, would be split into sub-tokens (see Table~\ref{tab:token-split}).

\begin{table}[!ht]
\centering
\Huge
\caption{Example of key political terms and their encoding in BERT and RooseBERT vocabularies.}
\resizebox{\columnwidth}{!}{%
\begin{tabular}{lll}
\toprule
\textbf{Word} & \textbf{\begin{tabular}[c]{@{}c@{}}BERT\\ vocabulary\end{tabular}} & \textbf{\begin{tabular}[c]{@{}c@{}}RooseBERT\\ vocabulary\end{tabular}} \\ \midrule
annuity       & ['an', '\#\#nu', '\#\#ity']              & annuity                \\
bureaucrat    & ['bureau', '\#\#crat']                   & bureaucrat             \\
deterrent     & ['de', '\#\#ter', '\#\#rent']            & deterrent              \\
undermining   & ['under', '\#\#mini', '\#\#ng']          & undermining            \\
endorse       & ['end', '\#\#ors', '\#\#e']              & endorse                \\
statutorily   & ['s', '\#\#tat', '\#\#uto', '\#\#rily']  & statutorily            \\
consequential & ['con', '\#\#se', '\#\#quential']        & consequential          \\
importer      & ['import', '\#\#er']                     & importer               \\ \bottomrule
\end{tabular}%
}
\label{tab:token-split}
\end{table}

\subsection{Evaluation Tasks}
To evaluate the performance of RooseBERT, we selected six downstream tasks: Sentiment Analysis, Stance Detection, Argument Component Detection and Classification, Argument Relation Prediction and Classification, Motion Policy Classification and NER. These tasks were tested using ten datasets in different settings (single-sentence classification, sentence-pair classification, multi-class classification and cross-domain). These tasks are central to NLP research on political debates, and publicly available annotated datasets enable systematic model evaluation. Accuracy was used as the evaluation metric for Sentiment Analysis and Stance Detection, while Average Macro F1 was employed for the remaining tasks, in line with the metrics adopted by the original dataset authors.

\subsubsection{Sentiment Analysis}
Sentiment analysis aims to determine whether the polarity of a text is positive or negative.

\noindent \textbf{ParlVote}~\cite{abercrombie2020parlvote} is a large annotated corpus of parliamentary debates. It consists of 34,010 UK parliamentary debates, 
labelled at the speech level (17,721 positive and 15,740 negative). We formulate this task as a sentence-pair classification problem, where the model predicts speech sentiment from the concatenated motion and speech text.

\noindent \textbf{HanDeSet}~\cite{abercrombie-batista-navarro-2018-aye} is a corpus of annotated parliamentary debates in the UK. It consists of 1,251 units, each composed of a parliamentary speech of up to five utterances and an associated motion. We cast this task as a single sentence classification problem where each speech has two binary sentiment labels.

\subsubsection{Stance Detection}
Stance detection aims to classify the attitude of a text towards a given target as \textit{support} or \textit{oppose}.

\noindent \textbf{ConVote}~\cite{Thomas2006ConVote} is an annotated corpus of U.S. floor debates in the House of Representatives for the year 2005. The corpus consists of 53 debates and 3857 speech segments. We cast this task as a single-sentence classification problem.

\noindent \textbf{AusHansard}~\cite{Ng2025AusHansard} is an annotated corpus of Australian parliamentary debates, of 778 speech segments. The authors use this to evaluate cross-domain adaptation by training and validating on augmented ParlVote and HanDeSeT, and testing on AusHansard. We follow the same setup to assess our model’s cross-domain adaptability.

\subsubsection{Argument Component Detection and Classification} 
In \textit{Argument Mining} (AM), detecting and classifying argument components is challenging, as it requires extracting argumentative units from natural language text, identifying their boundaries and classifying them as either \textit{premises} or \textit{claims}~\cite{lawrence2019complinguistic}.

\noindent \textbf{ElecDeb60to20}~\cite{goffredo2023emnlp} is a collection of 44 televised U.S. presidential debates from 1960 to 2020. The dataset comprises 38,667 annotated argument components (25,078 claims and 13,589 premises). We formulate this task as a sequence labelling problem using the BIO scheme at the speech-turn level, where each speech turn consists of multiple sentences of the same speaker.

\noindent \textbf{ArgUNSC}~\cite{PoiaganovaS25} is a collection of 144 speeches delivered by representatives of 24 different nations at the United Nations Security Council. The dataset includes 4,604 annotated argument components (2,348 claims and 2,256 premises). We formulate this task as a sequence labelling problem using the BIO scheme at the sentence level.

\subsubsection{Argument Relation Prediction and Classification}
In AM, argumentative components, i.e., \textit{premises} and \textit{claims}, can either \textit{support} or \textit{attack} other argument components, forming complex relational structures~\cite{lawrence2019complinguistic}. 

\noindent \textbf{ElecDeb60to20} provides annotations for argument relations (21,689 support, 3,835 attack). The task is formulated as multi-class sentence-pair classification: \textit{support}, \textit{attack}, \textit{no-relation}.

\noindent \textbf{ArgUNSC} also provides annotations for argument relations (2,623 support, 350 attacks). The task is formulated as multi-class sentence-pair classification: \textit{support}, \textit{attack}, \textit{no-relation}.

\subsubsection{Motion Policy Classification}
Policy preferences are commonly used in political science to describe politicians’ positions. The Manifesto Project~\footnote{\url{https://manifestoproject.wzb.eu}} defines a set of policy preference codes organised into seven domains. The coding scheme includes 74 policy preference codes, most of which are positional and indicate either a positive or negative stance toward a policy issue.

\noindent \textbf{ParlVote+}~\cite{abercrombie-batista-navarro-2022-policy} extends the ParlVote corpus, annotated with policy preferences. In total, the corpus contains 23,181 speeches annotated with 34 policy preferences. We cast this task as a single-sentence classification problem where the model has to classify the policy preference of the given speech.

\subsubsection{NER}
NER aims to locate and classify named entities in unstructured text into pre-defined categories.

\noindent \textbf{NEREx}, proposed by \cite{AssadySGKC17}, is an interactive visual analytics approach for exploring verbatim conversational transcripts. Their dataset of U.S. Presidential Debates contains 34 debates annotated with 37 entities. We evaluate this task at the sentence level. Although the dataset comes from a political domain, its annotated entities follow standard named entity categories and do not encode domain-specific political semantics.
\begin{table*}[!ht]
    \centering\caption{Perplexity evaluated on a held-out corpus representing the overall distribution of political debates. }
    \setlength{\tabcolsep}{3pt}
    \resizebox{\linewidth}{!}{%
        \begin{tabular}{@{}cccccccc@{}}
            \toprule
            \textit{\textbf{BERT's tokenizer}} & \textbf{BERT} & \textbf{ModernBERT} & \textbf{ConfliBERT-cont} & \textbf{ConfliBERT-scr} & \textbf{PoliBERTweet} & \textbf{RooseBERT-cont} & \textbf{RooseBERT-scr} \\ \midrule
            \textit{cased}     & 22.11 & 811048014.47 & 4.37 & 790959.65 & 255326.23 & 2.61 & 26320.05 \\
            \textit{uncased}   & 9.60 & N/A & 5.00 & 3270380.04 & N/A & 2.71 & 446395381.08 \\ \bottomrule
        \end{tabular}%
    }
    \label{tab:perplexity-base}
\end{table*}

\section{Experimental Setup}
\label{sec:experimental_setup}
In this section, we detail the design choice and technical elements necessary to train RooseBERT.
\subsection{Pre-training Setup}
The choice of BERT-base as backbone architecture is empirically motivated: we evaluated larger encoder architectures (e.g., ModernBERT) during preliminary experiments and consistently observed overfitting on the domain-specific corpus. Although 11GB is large by domain-specific standards, it remains an order of magnitude smaller than general-purpose pre-training corpora, and higher-capacity models were unable to generalise.
BERT-base proved the optimal balance between representational capacity and regularisation for this dataset size, consistent with the literature on domain-specific pre-training at BERT scale~\cite{beltagy-etal-2019-scibert, lee2020biobert, chalkidis2020LegalBERT}.

RooseBERT was trained with both \textsc{cont} and \textsc{scr} approaches, in both \textit{cased} and \textit{uncased} versions. Initially, the models were trained to optimise both the Masked Language Modeling (MLM), with a masking probability of 15\%, and the Next Sentence Prediction (NSP) objectives. However, as noted in several studies~\cite{liu2019roberta, joshi2020spanbert, hu-etal-2022-conflibert}, incorporating NSP alongside MLM did not lead to significant performance gains compared to training with MLM alone and it only slowed down the training. Therefore, we chose to discard the NSP objective.

Each pre-training dataset was split into 90\% training and 10\% testing, with the test set also reserved for perplexity evaluation. These splits were merged to form the final dataset, ensuring balanced representation across sources.

We experimented with various hyper-parameters to find the configuration that best minimised perplexity and loss on the test dataset. Specifically, we tested learning rates of $\{1, 3, 5\} \times e^{-4}$ and $\{2\} \times e^{-5}$. The learning rate was linearly warmed up over the first 10{,}000 steps, followed by a linear decay for the remainder of training. Once the learning rate was fixed, we experimented with varying number of training steps $\in$ \{50K , 100K , 150K , 200K , 250K , 300K\}.
Following~\cite{devlin-etal-2019-bert}, each model was trained for 80\% of the steps with a maximum sequence length of 128 and  with a maximum sequence length of 512 for the remaining 20\%.
For each training, we used eight A100 GPUs and the Transformer Library of HuggingFace with Python 3.10. Each training run was optimised using DeepSpeed ZeRO-2 methods, gradient accumulation, and FP16 precision to reduce training time. For training, we employed 8 devices with a per-device batch size of 64 with 4 gradient accumulation steps.

\subsection{Fine-tuning Setup}
Each PLM was fine-tuned on the downstream tasks described in Section~\ref{sec:methodology}.
Following \cite{devlin-etal-2019-bert}, we experimented with: learning rates of $\{2, 3, 5\} \times e^{-5}$; batch sizes of $8$, $16$, and $32$ and number of epochs of $2$, $3$, or $4$. For each model, we used its maximum available sequence length.
For all tasks, when available, we used the original train, dev and test splits, provided by the authors. When not available, we splitted the dataset into 80\% training, 10\% validation and 10\% testing.

RooseBERT's performances on downstream tasks were compared to the performances of BERT-base (used as a baseline), ModernBERT and of similar political PLMs, ConfliBERT and PoliBERTweet, each fine-tuned on the downstream tasks with the same hyper-parameters as RooseBERT. For each run, we used one A100 or H100 GPU and Python 3.10.
\section{Evaluation}
\begin{table*}[!ht]
	\centering
	\Huge
    \caption{Average $\pm$ standard deviation on downstream tasks. Binary tasks are evaluated with Accuracy; all other tasks use Macro F1. Statistical significance: ** $p<0.05$, * $p<0.1$ w.r.t. the best model (in \textbf{bold}). \textit{CC} = cont cased, \textit{CU} = cont uncased, \textit{SC} = scr cased, \textit{SU} = scr uncased, \textit{RooseB} = RooseBERT, \textit{ConfliB} = ConfliBERT.}
	\setlength{\tabcolsep}{3pt}
	\renewcommand{\arraystretch}{1.3}
	\resizebox{\linewidth}{!}{%
		\begin{tabular}{@{}lllllllllll@{}}
			\toprule
			\multirow{2}{*}{\textbf{Models}} &
			\multicolumn{2}{c}{\textbf{Sentiment Analysis}} &
			\multicolumn{2}{c}{\textbf{Stance Detection}} &
			\multicolumn{2}{c}{\textbf{\begin{tabular}[c]{@{}c@{}}
						                           Arg. Component \\ Det. \& Class.
			\end{tabular}}} &
			\multicolumn{2}{c}{\textbf{\begin{tabular}[c]{@{}c@{}}
						                           Arg. Relation \\ Pred. \& Class.
			\end{tabular}}} &
			\textbf{\begin{tabular}[c]{@{}c@{}}
						        Mot. Policy \\ Class.
			\end{tabular}} &
			\textbf{NER} \\
			\cmidrule(lr){2-3}\cmidrule(lr){4-5}\cmidrule(lr){6-7}\cmidrule(lr){8-9}\cmidrule(lr){10-10}\cmidrule(lr){11-11}
							& \textbf{ParlVote} 		& \textbf{HanDeSeT}			& \textbf{ConVote} 			& \textbf{AusHansard}		& \textbf{ElecDeb} 	& \textbf{ArgUNSC}			& \textbf{ElecDeb} 	& \textbf{ArgUNSC}			& \textbf{ParlVote+} 		& \textbf{NEREx} \\
			\midrule
			BERT-cased 		& 0.69 $\pm$ 0.02** 		& 0.67 $\pm$ 0.04** 		& 0.72 $\pm$ 0.01**			& 0.54 $\pm$ 0.01**			& 0.61 $\pm$ 0.00**			& 0.61 $\pm$ 0.01			& 0.58 $\pm$ 0.01**			& 0.57 $\pm$ 0.04**			& 0.54 $\pm$ 0.01**			& 0.92 $\pm$ 0.01\\
			BERT-uncased 	& 0.67 $\pm$ 0.08** 		& 0.66 $\pm$ 0.04** 		& 0.73 $\pm$ 0.01**			& 0.55 $\pm$ 0.02**			& 0.61 $\pm$ 0.00**			& 0.60 $\pm$ 0.01**			& 0.58 $\pm$ 0.01**			& 0.64 $\pm$ 0.03**			& 0.55 $\pm$ 0.01**			& 0.90 $\pm$ 0.01*\\
            ModernBERT 		& 0.77 $\pm$ 0.03 			& 0.65 $\pm$ 0.02** 		& 0.73 $\pm$ 0.04 			& 0.58 $\pm$ 0.05  			& 0.62 $\pm$ 0.00 			& 0.58 $\pm$ 0.03 			& 0.60 $\pm$ 0.01 			& 0.60 $\pm$ 0.14 			& \textbf{0.66 $\pm$ 0.00} 	& 0.91 $\pm$ 0.01\\
			ConfliB CC 		& 0.76 $\pm$ 0.01** 		& 0.65 $\pm$ 0.01**			& 0.75 $\pm$ 0.01**			& 0.61 $\pm$ 0.01**			& 0.62 $\pm$ 0.00*			& 0.61 $\pm$ 0.00*			& 0.61 $\pm$ 0.01**			& 0.55 $\pm$ 0.18*			& 0.58 $\pm$ 0.01**			& 0.90 $\pm$ 0.02*\\
			ConfliB CU 		& 0.75 $\pm$ 0.01**			& 0.65 $\pm$ 0.01**			& 0.75 $\pm$ 0.01**			& 0.57 $\pm$ 0.01**			& 0.62 $\pm$ 0.00**			& 0.60 $\pm$ 0.01**			& 0.61 $\pm$ 0.00**			& 0.62 $\pm$ 0.01**			& 0.55 $\pm$ 0.01**			& 0.91 $\pm$ 0.00**\\
			ConfliB SC 		& 0.74 $\pm$ 0.00** 		& 0.65 $\pm$ 0.01**			& 0.75 $\pm$ 0.01*			& 0.57 $\pm$ 0.02**			& 0.62 $\pm$ 0.00*			& 0.61 $\pm$ 0.01**			& 0.58 $\pm$ 0.02**			& 0.40 $\pm$ 0.23**		 	& 0.57 $\pm$ 0.00**			& \textbf{0.92 $\pm$ 0.00}\\
			ConfliB SU 		& 0.74 $\pm$ 0.02** 		& 0.64 $\pm$ 0.01**			& 0.75 $\pm$ 0.01**			& 0.59 $\pm$ 0.01**			& 0.62 $\pm$ 0.00**			& 0.60 $\pm$ 0.01**			& 0.35 $\pm$ 0.15**			& 0.69 $\pm$ 0.02**			& 0.59 $\pm$ 0.01**			& 0.91 $\pm$ 0.01\\
			PoliBERTweet 	& 0.58 $\pm$ 0.04** 		& 0.60 $\pm$ 0.09** 		& 0.68 $\pm$ 0.02**			& 0.54 $\pm$ 0.01**			& 0.60 $\pm$ 0.01**			& 0.61 $\pm$ 0.01**			& 0.54 $\pm$ 0.19			& 0.67 $\pm$ 0.02**			& 0.34 $\pm$ 0.19**			& 0.88 $\pm$ 0.01**\\
			RooseB CC 		& 0.79 $\pm$ 0.00** 		& \textbf{0.74 $\pm$ 0.03} 	& 0.76 $\pm$ 0.01			& \textbf{0.63 $\pm$ 0.02}	& \textbf{0.63 $\pm$ 0.01}	& \textbf{0.62 $\pm$ 0.00}	& 0.61 $\pm$ 0.00**			& 0.70 $\pm$ 0.02			& 0.62 $\pm$ 0.01**			& 0.90 $\pm$ 0.02\\
			RooseB CU 		& \textbf{0.80 $\pm$ 0.00} 	& 0.71 $\pm$ 0.04** 		& \textbf{0.77 $\pm$ 0.01}	& 0.60 $\pm$ 0.01**			& 0.62 $\pm$ 0.01			& 0.61 $\pm$ 0.00**			& 0.61 $\pm$ 0.01**			& 0.70 $\pm$ 0.02			& 0.63 $\pm$ 0.01**			& 0.90 $\pm$ 0.01**\\
			RooseB SC 		& 0.79 $\pm$ 0.00 			& 0.71 $\pm$ 0.02			& 0.75 $\pm$ 0.01**			& 0.60 $\pm$ 0.02*			& 0.63 $\pm$ 0.00			& 0.61 $\pm$ 0.00**			& \textbf{0.63 $\pm$ 0.00}	& \textbf{0.72 $\pm$ 0.01}	& 0.62 $\pm$ 0.00**		 	& 0.88 $\pm$ 0.02**\\
			RooseB SU 		& 0.78 $\pm$ 0.00** 		& 0.69 $\pm$ 0.02			& 0.75 $\pm$ 0.01			& 0.59 $\pm$ 0.01**			& 0.63 $\pm$ 0.01			& 0.61 $\pm$ 0.01			& 0.62 $\pm$ 0.01**			& 0.65 $\pm$ 0.05**			& 0.58 $\pm$ 0.01**			& 0.88 $\pm$ 0.01**\\
			\bottomrule
		\end{tabular}%
	}
	\label{tab:results-base}
\end{table*}
\label{sec:results}
In this section, we detail the results of the pre-training and fine-tuning processes.
\subsection{Pre-Training}
Perplexity (PPL)~\cite{perplexity} was used as a metric to determine the best performing pre-trained model, given the hyper-parameters previously described. It is defined as the exponentiated average negative log-likelihood of a sequence $X~=~(x_0, x_1, x_2 .. x_t)$ as follows: 
\begin{equation}
    PPL(X) = exp \{ - \frac{1}{t} \sum^{t}_{i}\ log\ p_\theta(x_i | x_{<i}) \}
\end{equation}

Perplexity was used to evaluate the model’s ability to predict masked tokens on an unseen test dataset representative of the overall corpus distribution. A lower perplexity indicates that the model assigns higher probabilities to the correct next words, i.e., it is more confident in its predictions, less \textit{confused} by the text. For a fair comparison, all models in this task used the original BERT tokenizer, with either the cased or uncased variant selected according to the specific model configuration.

Table~\ref{tab:perplexity-base} reports the perplexity scores of each model. For conciseness, we only report the best-performing configurations for RooseBERT.
Specifically, for both RooseBERT-cont and RooseBERT-scr, a learning rate of  $3e^{-4}$  consistently yielded the lowest perplexity on the test dataset (see Section~\ref{sec:methodology}). The total number of steps for the \textsc{cont} model was fixed to 150K and 250K for the \textsc{scr} model.


As shown in Table~\ref{tab:perplexity-base}, both BERT and ModernBERT exhibits a high perplexity, underscoring the challenge that general-purpose models face in capturing the distribution of domain-specific data without further adaptation. In contrast, ConfliBERT achieves lower perplexity scores, with ConfliBERT-cont yielding results most closely aligned with RooseBERT-cont. Among all models, RooseBERT-cont demonstrates the lowest perplexity, indicating a more effective adaptation to the political debate domain.

Interestingly, the \texttt{scr} versions of ConfliBERT and RooseBERT, as well as PoliBERTweet, fail to converge when using BERT's vocabulary, likely due to the mismatch between their domain-specific tokenization and the original BERT tokenizer. ModernBERT similarly exhibits unexpectedly high perplexity, further suggesting that its tokenization scheme is fundamentally incompatible with BERT's vocabulary.

Concerning training time, \textsc{cont} models required 24 hours and \textsc{scr} 33 hours, both using 8 A100 GPUs. Fine-tuning requires only a single A100 or H100 GPU. This stands in contrast to ModernBERT, whose pre-training at scale requires substantially greater resources. Despite this compute gap, RooseBERT achieves competitive or superior downstream performance, demonstrating that targeted domain adaptation on a BERT-scale architecture is an efficient alternative to scaling up general-purpose models.

\subsection{Fine-Tuning}
BERT, ModernBERT, RooseBERT, ConfliBERT and PoliBERTweet available versions (\textsc{cont}, \textsc{scr}, \textit{cased} and \textit{uncased}) were fine-tuned on the downstream tasks using the hyper-parameters of Section \ref{sec:experimental_setup}. In particular, ConfliBERT is available in both \textit{cased} and \textit{uncased} version, with \textsc{cont} and \textsc{scr} configurations. PoliBERTweet is only available in \textit{cased} \textsc{cont} configuration and ModernBERT is only available in \textit{cased} configuration.

Table \ref{tab:results-base} shows fine-tuning results. Each score represents the average performance over 5 independent runs, initialised with different seeds (42, 12, 48, 16 and 33). RooseBERT achieves better performance on 8 out of 10 tasks. Notably, this includes competitive results over ModernBERT despite RooseBERT's substantially lower training cost and simpler BERT-base architecture.

In the binary classification tasks, RooseBERT yields statistically significant improvements over the baseline model (BERT) ranging from 4\% on ConVote to 11\% on ParlVote. These results indicate a consistent benefit from domain-specific pre-training.

For argument detection and classification, RooseBERT achieves statistically significant improvements on the ElecDeb60to20 dataset, with gains of 2\% over the baseline. ArgUNSC is a harder benchmark for all models: performance across systems clusters tightly in the 0.60–0.62 range, a compressed space that reduces statistical power even when genuine differences exist. Nonetheless, RooseBERT achieves the highest mean score, and the ablation study (Table~\ref{tab:results-ablation}) confirms that debate-structural features are learned even without UN-specific pre-training data, ruling out data leakage as the source of the advantage.

In the argument relation prediction task, RooseBERT improves over the baseline on both datasets. On ElecDeb60to20, the comparison with PoliBERTweet is not statistically significant, but this reflects PoliBERTweet's instability rather than genuine parity: its standard deviation of 0.19 on this task indicates severe sensitivity to random seed initialisation. In contrast, RooseBERT's standard deviation remains consistently below 0.02 across all seeds, reflecting more reliable and robust training.

For motion policy classification, RooseBERT shows statistically significant gains, outperforming the baseline by 8\% and other models by 4\%. The exception is ModernBERT, which achieves the highest score on this task (0.66 vs.\ RooseBERT's 0.63). This appears to be the one task where ModernBERT's broader representational capacity provides an advantage, likely due to the diversity of policy labels requiring wide semantic coverage. 

In NER, ConfliBERT achieves the highest score, though not significantly better than BERT or RooseBERT. More importantly, since the NER dataset uses general rather than politically-specific entity categories, domain-adaptive pre-training provides no structural advantage, and none is expected. It is important to highlight that RooseBERT does not degrade on general tasks: it maintains performance comparable to the BERT baseline, demonstrating an absence of negative transfer. This makes RooseBERT a valuable replacement for BERT in mixed NLP pipelines that combine political and general data.

Overall, RooseBERT achieves best performance on 8 out of 10 tasks. It outperforms the other models in single-output classification tasks (e.g., sentiment analysis, stance detection) and maintains competitive performance as task complexity increases. Crucially, it matches or outperforms ModernBERT across most tasks while using a simpler architecture trained at significantly lower cost.

ElecDeb60to20 and ArgUNSC represent difficult benchmarks for all models, with lower scores and limited performance separation. However, results indicate that domain-adaptive pre-training is beneficial and RooseBERT's gains justify its use for complex argumentation tasks.

We also experimented with Large Language Models (LLMs). Our findings show that while zero and few-shot methods offer a practical alternative to fine-tuning, LLMs perform competitively mainly when fine-tuned. Performance drops as task complexity increases, especially for argument component detection and classification, while simple tasks like sentiment analysis remain more stable. Overall, LLM performance relies heavily on extensive training, whereas RooseBERT offers a lightweight alternative with strong results.

\subsection{Ablation Study}
\begin{figure}[!ht]
    \centering
    \includegraphics[width=\linewidth]{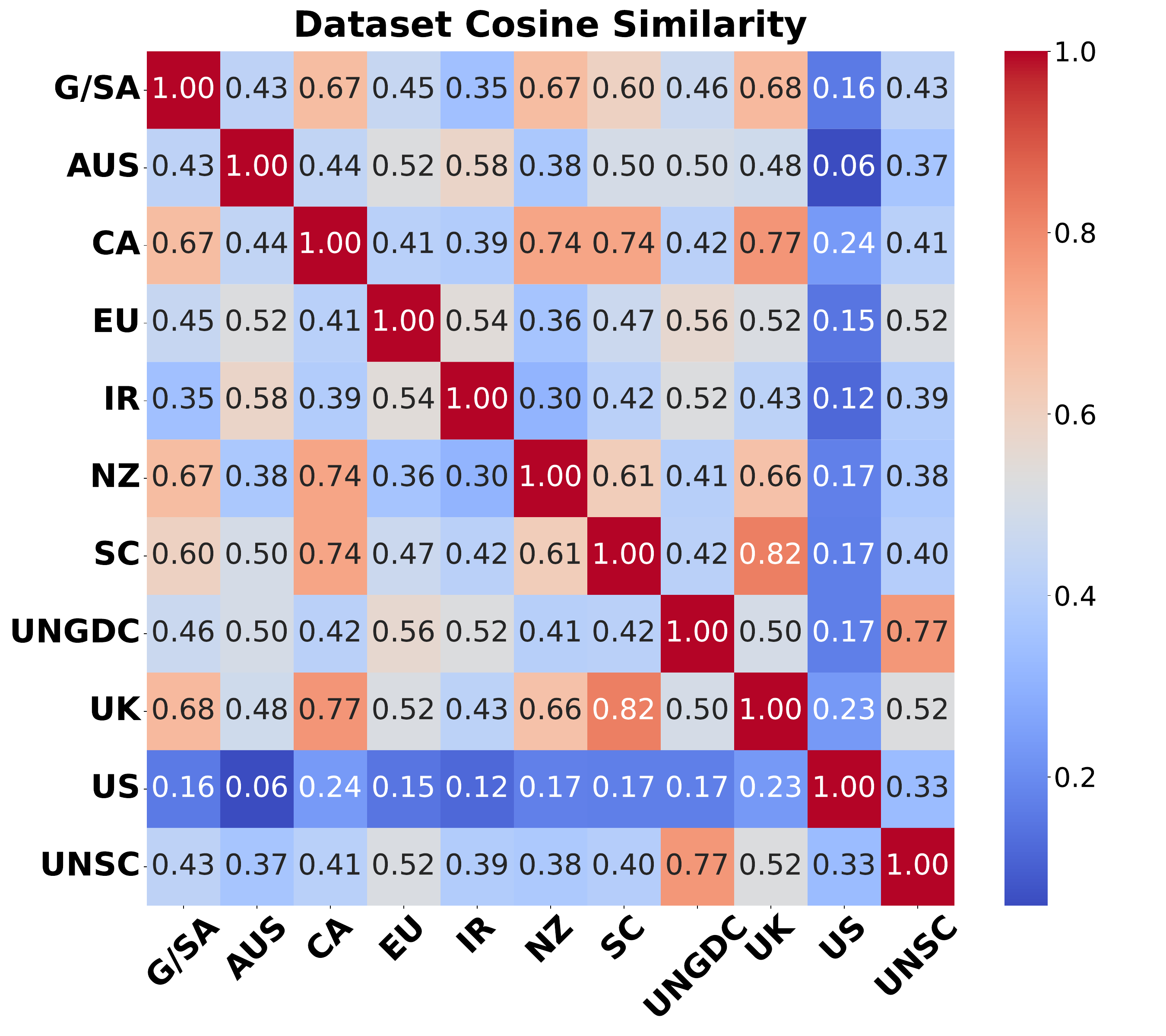}
    \caption{Cosine similarity among pre-training datasets. Dataset abbreviations: G/SA (Ghana and South Africa Parliamentary debates), AUS (Australian Parliamentary debates), CA (Canadian Parliamentary debates), EU (European Parliamentary debates), IR (Irish Parliamentary debates), NZ (New Zealand Parliamentary debates), SC (Scottish Parliamentary debates), UNGDC (United Nations General Debate Corpus), UK (United Kingdom Parliamentary debates), US (United States Debates), and UNSC (United Nations Security Council).}
    \label{fig:similarity}
\end{figure}

\begin{table*}[!ht]
    \centering
    \Huge
    \caption{Average $\pm$ standard deviation on downstream tasks. Binary tasks are evaluated with Accuracy; all other tasks use Macro F1. Statistical significance: ** $p<0.05$, * $p<0.1$ w.r.t. to RB SU (RooseBERT-scr-uncased).}
    \setlength{\tabcolsep}{3pt}
    \renewcommand{\arraystretch}{1.3}
    \resizebox{\linewidth}{!}{%
        \begin{tabular}{@{}lllllllllll@{}}
            \toprule
            \multirow{2}{*}{\textbf{Clusters}} &
            \multicolumn{2}{c}{\textbf{Sentiment Analysis}} &
            \multicolumn{2}{c}{\textbf{Stance Detection}} &
            \multicolumn{2}{c}{\textbf{\begin{tabular}[c]{@{}c@{}}Arg. Component \\ Det. \& Class.\end{tabular}}} &
            \multicolumn{2}{c}{\textbf{\begin{tabular}[c]{@{}c@{}}Arg. Relation \\ Pred. \& Class.\end{tabular}}} &
            \textbf{\begin{tabular}[c]{@{}c@{}}Mot. Policy \\ Class.\end{tabular}} &
            \textbf{NER} \\
            \cmidrule(lr){2-3}\cmidrule(lr){4-5}\cmidrule(lr){6-7}\cmidrule(lr){8-9}\cmidrule(lr){10-10}\cmidrule(lr){11-11}
                        & \textbf{ParlVote}     & \textbf{HanDeSeT} & \textbf{ConVote}  & \textbf{AusHansard}   & \textbf{ElecDeb}    & \textbf{ArgUNSC}  & \textbf{ElecDeb}    & \textbf{ArgUNSC}      & \textbf{ParlVote+}    & \textbf{NEREx} \\
            \midrule
            \#0  & 0.79 $\pm$ 0.00**    & 0.69 $\pm$ 0.03   & 0.76 $\pm$ 0.01   & 0.62 $\pm$ 0.01**    & 0.63 $\pm$ 0.00           & 0.61 $\pm$ 0.01   & 0.63 $\pm$ 0.00**         & 0.70 $\pm$ 0.01**     & 0.63 $\pm$ 0.01**    & 0.89 $\pm$ 0.01\\
            \#1  & 0.79 $\pm$ 0.01**     & 0.73 $\pm$ 0.02*  & 0.78 $\pm$ 0.00**& 0.62 $\pm$ 0.01**     & 0.62 $\pm$ 0.00           & 0.61 $\pm$ 0.01   & 0.63 $\pm$ 0.01**        & 0.72 $\pm$ 0.02*      & 0.62 $\pm$ 0.00**    & 0.88 $\pm$ 0.01\\
            \#2  & 0.78 $\pm$ 0.00       & 0.67 $\pm$ 0.03   & 0.76 $\pm$ 0.01   & 0.59 $\pm$ 0.01       & 0.62 $\pm$ 0.00*          & 0.60 $\pm$ 0.01*  & 0.62 $\pm$ 0.00**         & 0.71 $\pm$ 0.03*      & 0.60 $\pm$ 0.01**     & 0.90 $\pm$ 0.02\\
            \#3  & 0.69 $\pm$ 0.09*      & 0.70 $\pm$ 0.04   & 0.72 $\pm$ 0.02** & 0.57 $\pm$ 0.02*      & 0.52 $\pm$ 0.01**        & 0.59 $\pm$ 0.01**& 0.61 $\pm$ 0.00**         & 0.71 $\pm$ 0.03*      & 0.60 $\pm$ 0.01**     & 0.91 $\pm$ 0.02\\
            \textbf{RB SU}       & \textbf{0.78 $\pm$ 0.00}       & \textbf{0.69 $\pm$ 0.02}   & \textbf{0.75 $\pm$ 0.01}   & \textbf{0.59 $\pm$ 0.01}       & \textbf{0.63 $\pm$ 0.01}           & \textbf{0.61 $\pm$ 0.01}   & \textbf{0.62 $\pm$ 0.01}           & \textbf{0.65 $\pm$ 0.05}       & \textbf{0.58 $\pm$ 0.01}       & \textbf{0.88 $\pm$ 0.01}\\
            \bottomrule
        \end{tabular}%
    }
    \label{tab:results-ablation}
\end{table*}
To identify which parts of the dataset most influence our results, we conducted an ablation study. Since the dataset has eleven subsets, a full leave-one-out ablation was computationally expensive. Instead, we clustered the subsets based on content similarity, using cosine similarity over the average sentence embeddings computed with \texttt{all-mpnet-base-v2}\footnote{\url{https://huggingface.co/sentence-transformers/all-mpnet-base-v2}} Sentence Transformer.
Figure~\ref{fig:similarity} shows the clustering results obtained using Agglomerative Clustering with a distance threshold of 0.5. 

Four distinct clusters emerge. \textit{Cluster\#0} contains the European Debates, UNSC and UNGDC reflecting their shared institutional origin and likely overlap in terminology and thematic content, \textit{Cluster\#1} contains the Australian and Irish Debates, \textit{Cluster\#2} contains debates from UK, Scotland, Africa, Canada and New Zealand, all Commonwealth countries, and \textit{Cluster\#3} only contains the debates from US, which is expected since they are the only ones not held in a parliamentary setting.

After clustering, we pre-trained four separate RooseBERT models, each using a leave-one-out strategy w.r.t. the clusters.
For instance, the model labelled \textit{Cluster\#1} was trained on all debates except those in Cluster 1 (i.e., Australian and Irish debates). 
This setup enables us to evaluate each cluster by measuring the effect of its exclusion.

For model training, the \textsc{scr} configuration was chosen over \textsc{cont} to evaluate the debates in isolation, avoiding influence from prior training on the general corpus. The \textit{uncased} variant was used for efficiency, requiring less computational time than the cased version. Hyperparameters were set according to RooseBERT-scr-uncased.


Table~\ref{tab:results-ablation} shows the results of the fine-tuning of these models on the four different downstream tasks. Results are averaged on a total of 5 runs with different seeds (42, 12, 48, 16, 33).

Models trained with different dataset configurations remain largely consistent across all downstream tasks, with minimal to no degradation compared to the original RooseBERT model.
This consistency underscores RooseBERT's strong generalisation capabilities regardless of the training data subset used.
Notably, \textit{Cluster\#0}, trained on United Nations datasets, shows no performance drop on ArgUNSC tasks, and even improves argument relation classification.
\textit{Cluster\#1}, trained on Australian parliamentary debates, maintains overall performance and improves stance detection on AusHansard.
\textit{Cluster\#2}, trained on UK debates, remains stable across ParlVote and HanDeSeT tasks, further demonstrating robust domain-specific fine-tuning.
Finally, \textit{Cluster\#3}, trained on US debates, shows a moderate performance drop in argument component detection and classification, but remains considerably stable on argument relation prediction and classification.

The results show no significant decrease, suggesting that the structural patterns for these tasks can be effectively learned from other debates. One might argue that RooseBERT's performance on downstream tasks is biased by test data overlapping with the large corpora used for PLM pre-training. However, Table~\ref{tab:results-ablation} shows the opposite: the ablation study demonstrates that the model can generalize the discourse structure of political debates even when a key domain is withheld during pre-training.
\section{Conclusion}
\label{sec:conclusion}
We present RooseBERT, a domain-specific pre-trained language model for English political debates, together with the largest publicly available English political debate corpus to date (11GB, spanning 1946–2025 across 11 geopolitical contexts. This corpus is released as a standalone contribution to the community, enabling future research in political NLP. RooseBERT is developed via two strategies: continued pre-training of BERT and training from scratch, in both \textit{cased} and \textit{uncased} variants. Evaluation on ten datasets across six downstream tasks shows that RooseBERT achieves best mean performance on 8 out of 10 datasets.

A key finding is that domain-adaptive pre-training on a BERT-scale architecture is a highly cost-effective strategy: RooseBERT matches or outperforms ModernBERT, a larger, more recent general-purpose model, across most tasks, while requiring only 24 hours of pre-training on 8 A100 GPUs and fine-tuning on a single GPU. This makes RooseBERT a cost-efficient choice for NLP tasks involving political data.

To the best of our knowledge, RooseBERT is the first language model tailored specifically for political debates.
With the increasing influence of political discourse on public opinion and the spread of misinformation, developing accessible domain-specific models is both timely and essential.

Future work will expand RooseBERT to other languages, enabling cross-lingual analysis of political discourse and comparative studies across different political argumentation styles.

\section*{Acknowledgment}
This work was supported by the French government through the 3IA Côte d'Azur Investments (ANR-23-IACL-0001) and the France 2030 investment plan as part of the Initiative of Excellence Université Côte d'Azur (ANR-15-IDEX-01) and by the ANR project ATTENTION (ANR-21-CE23-0037). It is also supported by ERC grant PANDORA – 101231869. Funded by the European Union. Views and opinions expressed are however those of the author(s) only and do not necessarily reflect those of the European Union or the European Research Council Executive Agency. Neither the European Union nor the granting authority can be held responsible for them. Neither can be held responsible for them. This project was provided with computing AI and storage resources by GENCI at IDRIS thanks to the grant 2026-AD011016047R1 on the supercomputer Jean Zay’s A100 partition. The authors thank Franck Diard for providing computing resources on the DR-1 GPU cluster.

\bibliographystyle{IEEEtran}
\bibliography{custom}
\end{document}